\documentclass[10pt,twocolumn,letterpaper]{article}

\usepackage[pagenumbers]{cvpr} % To force page numbers, e.g. for an arXiv version

\usepackage{graphicx}
\usepackage{amsmath}
\usepackage{amssymb}
\usepackage{booktabs}

\usepackage{multirow}
\usepackage{multicol}
\usepackage{makecell}
\usepackage{bbding}

\usepackage[pagebackref,breaklinks,colorlinks]{hyperref}

\usepackage[capitalize]{cleveref}
\crefname{section}{Sec.}{Secs.}
\Crefname{section}{Section}{Sections}
\Crefname{table}{Table}{Tables}
\crefname{table}{Tab.}{Tabs.}

\def\cvprPaperID{7898} % *** Enter the CVPR Paper ID here
\def\confName{CVPR}
\def\confYear{2023}

\begin{document}

\title{MonoSIM: Simulating Learning Behaviors of Heterogeneous Point Cloud Object Detectors for Monocular 3D Object Detection}

\author{Han Sun$^1$, Zhaoxin Fan$^2$\thanks{Equal contribution.}, Zhenbo Song$^1$, Zhicheng Wang$^3$, Kejian Wu$^3$, Jianfeng Lu$^{1}$\thanks{Corresponding author.}
\\
$^1$Nanjing University of Science and Technology
\\
$^2$Renmin University of China; $^3$Nreal
\\
}

% \and
\maketitle

%%%%%%%%%%%%%%%%%%%%%%%%%%%%%%%%%%%%%%%%%%%%%%%%%%%%%%%%%%%%%%%%%%%%%%%%%%%%%%%%%%%%%%%%%%%%%%%%%%%%%% ABSTRACT
\begin{abstract}
Monocular 3D object detection is a fundamental but very important task to many applications including autonomous driving, robotic grasping and augmented reality. Existing leading methods tend to estimate the depth of the input image first, and detect the 3D object based on point cloud. This routine suffers from the inherent gap between depth estimation and object detection. Besides, the prediction error accumulation would also affect the performance. In this paper, a novel method named MonoSIM is proposed. The insight behind introducing MonoSIM is that we propose to simulate the feature learning behaviors of a point cloud based detector for monocular detector during the training period. Hence, during inference period, the learned features and prediction would be similar to the point cloud based detector as possible. To achieve it, we propose one scene-level simulation module, one RoI-level simulation module and one response-level simulation module, which are progressively used for the detector's full feature learning and prediction pipeline. We apply our method to the famous M3D-RPN detector and CaDDN detector, conducting extensive experiments on KITTI and Waymo Open datasets. Results show that our method consistently improves the performance of different monocular detectors for a large margin without changing their network architectures. Our codes will be publicly available at \href{https://github.com/sunh18/MonoSIM}{https://github.com/sunh18/MonoSIM}.
\end{abstract}

%%%%%%%%%%%%%%%%%%%%%%%%%%%%%%%%%%%%%%%%%%%%%%%%%%%%%%%%%%%%%%%%%%%%%%%%%%%%%%%%%%%%%%%%%%%%%%%%%%%%%%% BODY TEXT
%%%%%%%%%%%%%%%%%%%%%%%%%%%%%%%%%%%%%%%%%%%%%%%%%%%%%%%%%%%%%%%%%%%%%%%%%%%%%%%%%%%%%%%%%%%%%%%%%%%%%%%
\section{Introduction}

\begin{figure}[t]
\centering
\setlength{\belowcaptionskip}{-0.4cm}
\includegraphics[scale=0.55]{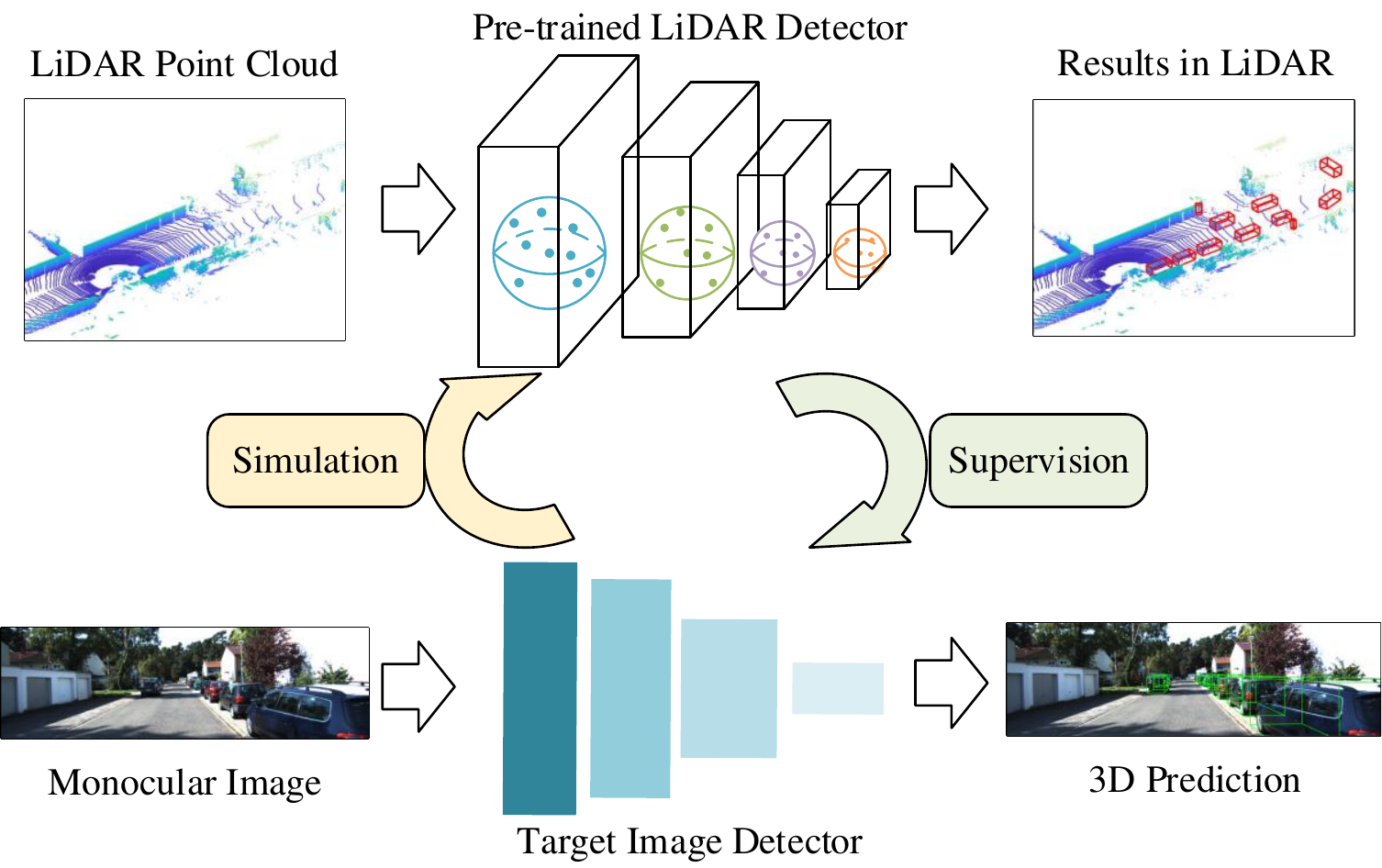}
\caption{The monocular detector is supervised by the point cloud based detector while directly simulating the feature learning behaviors, without extra depth estimation and changing the original network structures.}
\label{figure1:Monocular image network simulate point cloud network.}
\end{figure}

3D scene perception is an important component in many artificial intelligence scenarios, such as autonomous driving, robotics and augmented reality. Previous algorithms and solutions based on LiDAR\cite{lang2019pointpillars,shi2019pointrcnn,shi2020pv,shi2020points,zheng2021se} or stereo vision\cite{li2019stereo,chen2020dsgn} have achieved satisfactory detection performance. However, the high cost and installation requirements limit the wide application of these methods. Therefore, the cheaper and easy-to-deploy monocular 3D detection methods\cite{brazil2019m3d,luo2021m3dssd,weng2019monocular,zhang2021objects} become alternative solutions and show great potential. In this paper, we research monocular 3D object detection.

Existing monocular 3D detection methods can be roughly divided into direct methods \cite{mousavian20173d,li2019gs3d,brazil2019m3d,liu2019deep,manhardt2019RoI,qin2019monogrnet,simonelli2019disentangling}  and pseudo-LiDAR based methods methods \cite{wang2019pseudo,weng2019monocular,qian2020end,ma2020rethinking,ye2020monocular,ma2019accurate,reading2021categorical,fan2021deep}. The former directly predict 3D bounding box from an image, while the later estimate a dense depth map from the image first, then adopt a point cloud based 3D object detector to make usage of the estimated depth. Comparing the two kinds of methods,  pseudo-LiDAR based methods tend to achieve better performance since the recovered dense depth maps provide more cues for 3D geometric and semantic perception. Nevertheless, though pseudo-LiDAR based methods have achieved acceptable performance,  still face several certain disadvantages: 1)  The depth estimation and 3D object detection are achieved by two different deep networks. The supervision signal for the 3D object detection cannot flow back to the depth estimation network to guide its training. Therefore, there is a gap between the two different tasks. 2) Since both two tasks are easy to fall into local optimal solution, the error of the two tasks would accumulate. Hence, the prior information hidden in the ground-truth depth information cannot be made full usage. 3) Using two different networks would increase computational overhead during both training and testing. \label{disadvantages of depth estimation}

To tackle above issue, we propose a novel monocular 3D object detection training pipeline named MonoSIM. The design insight behind MonoSIM is that we  believe the basic principle of pseudo-LiDAR based methods is to simulate the point cloud based 3D object detectors. Specifically,  we find most existing pseudo-LiDAR  choose to estimate the depth map and project it into point cloud. This process, in essence, is trying to simulating the input of the 3D object detector. This motivates us to think about that: Can we simulate the 3D object detector in a more straight-forward way? 

The answer is positive.  As shown in \cref{figure1:Monocular image network simulate point cloud network.}, in  MonoSIM, we propose to simulate the feature learning behavior of the 3D object detector instead of simulating its input.   To simulate the full feature learning of point cloud based pipelines, MonoSIM consists of three modules: 
1) Scene-Level Simulation Module, which aligns the shallow scene-level features of the point cloud detector and monocular detector, aims at increasing the monocular detector environmental understanding ability. 
2) RoI-Level Simulation Module, which simulates the feature characteristics of point cloud RoIs for each monocular RoI, targets at increasing the monocular detector's ability of finding and locating the potential objects.
3) Response-Level Simulation Module, which uses prediction of the point cloud based detector as soft labels to guide the loss computing step, aims at increasing the monocular detector's ability of regressing geometric properties of the bounding box. 
The above three module would progressively help our model inherit the strong  detection power from the point cloud based detector.  The simulation is very similar to knowledge distillation \cite{hinton2015distilling}, the difference lies that our work simulation idea can support cross modal information transmission (from point cloud modality to the monocular image modality).

Since our method is model agnostic, it can be applied to many different monocular 3D object detectors. In our work, we apply MonoSIM to the famous M3D-RPN detector and CaDDN detector. Extensive experiments are conduct on KITTI, the currently most widely used benchmark, and Waymo Open dataset, the currently most large scale dataset. Experimental results show that our method consistently improves the performance of different monocular detectors for a large margin without changing their network architectures on both datasets.
% And our method  achieves state-of-the-art performance on the KITTI benchmark.

Our contributions can be summarised as:
\begin{itemize}
    \item We propose MonoSIM, a novel pipeline that enables monocular 3D object detection method simulate the feature learning process of point cloud based detectors in scene-level, RoI-level and response-level, respectively.
    
    \item We apply our MonoSIM to several different existing monocular 3D object detectors. Our method consistently and significantly improves their performance without change their original network architectures.

    \item  We conduct extensive experiments on the most widely used KITTI benchmark and the most large scale Waymo Open dataset to verify the effectiveness of our method.
    
\end{itemize}

\begin{figure*}[t]
\centering
\setlength{\belowcaptionskip}{-0.4cm}
\includegraphics[width=1\textwidth]{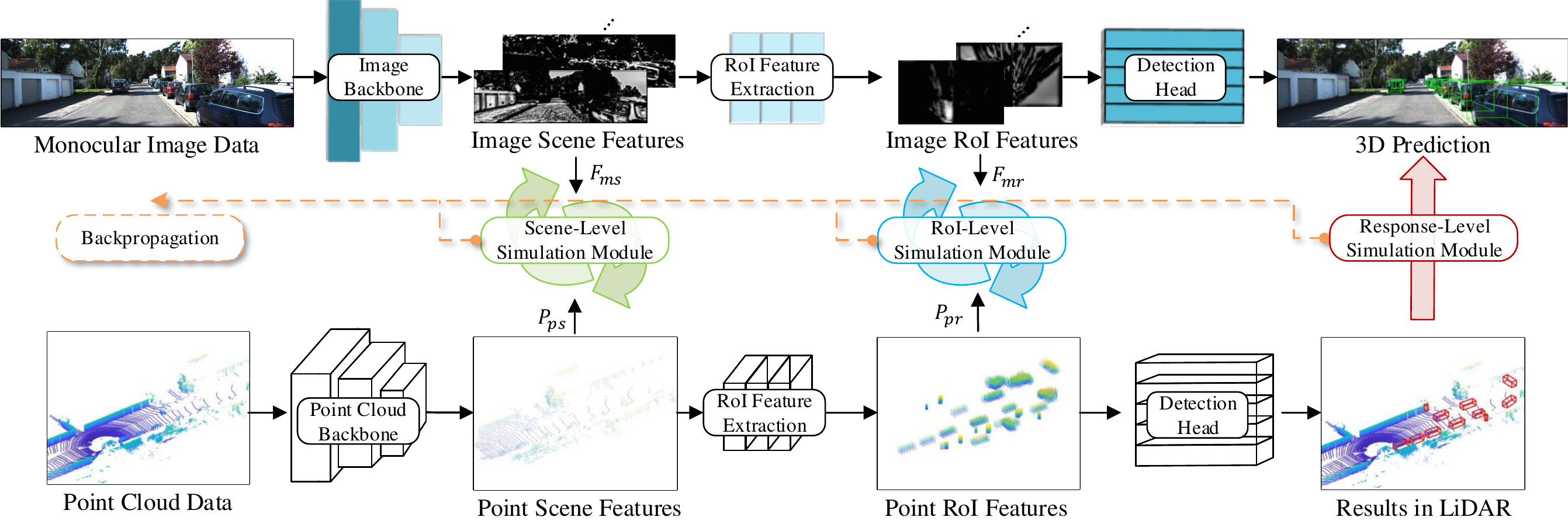}
\caption{The overall architecture of our proposed MonoSIM. First we divide the monocular detector and point cloud based detector into scene-level, RoI-level and response-level components to prepare for cross modal simulation. 
Then scene-level features of the two detectors are sent to the scene-level simulation module and RoI-level features are processed by the RoI-level simulation module. In the response-level simulation module, the training of monocular detector is supervised by the prediction of the point cloud based detector.
Finally, the above three levels of simulations are combined to form a complete pipeline to optimize the performance of the monocular detector.}
\label{figure2:Pipeline of MonoSIM}
\end{figure*}

%%%%%%%%%%%%%%%%%%%%%%%%%%%%%%%%%%%%%%%%%%%%%%%%%%%%%%%%%%%%%%%%%%%%%%%%%%%%%%%%%%%%%%%%%%%%%%%%%%%%%%%%
\section{Related Work}
\subsection{Point Cloud based 3D Object Detection}
Point cloud based 3D object detection methods can roughly divided into point-based methods and grid-based methods \cite{shi2020pv}. 
Point-based methods are represented by PointNet \cite{qi2017pointnet} and PointNet++ \cite{qi2017pointnet++}, which directly extract features on the raw point cloud via deep networks. On this basis, PointRCNN \cite{shi2019pointrcnn} applies PointNET++ as the backbone to explore the accurate location of 3D proposals. Point-based methods have the problems of high time cost and large amount of calculation, and some solutions such as AVOD \cite{Ku2018Joint} and F-PointNet \cite{qi2018frustum} are developed to reduce these effects, which use LiDAR point cloud and RGB images simultaneously. The former converts the point cloud into BEV (Bird's Eye View) maps, then the feature maps of RGB images and BEV maps are obtained by FPN network, and finally generates proposals after fusing the two types of features. The latter adopts 2D region proposal to guide 3D instance segmentation, reducing point cloud search. However, the quality of 2D detection will have an uncertain impact on 3D detection. Then, in grid-based methods, the point cloud is projected to regular grids \cite{lang2019pointpillars,yang2018hdnet} or divided into voxels \cite{zhou2018voxelnet}, and then these divided points are sent to the full connection layer to construct a unified feature representation. Finally, the features are extracted by 2D or 3D CNN for prediction. These grid-based methods are generally efficient for accurate 3D proposal generation, but the receptive fields are constraint by the kernel size of 2D/3D convolutions. Due to the sparse characteristics of point cloud, the above methods often need to combine sparse convolution \cite{graham20183d} or densification strategy \cite{yang2019std} to enhance feature representation. 

% Though point cloud based 3D object detection model have achieve excellent performance, the cost of collecting point cloud is expensive, making it unaffordable to many applications. In this paper, we research the much cheaper and flexible monocular 3D object detection.

\subsection{Monocular 3D Object Detection}

Monocular 3D object detection is a challenge task since recovering precise 3D information from a single RGB image is an ill-posed problem.  A straight forward line of works propose to directly predict 3D objects from the image. For example, Mono3D \cite{chen2016monocular} generates 3D anchors via semantic segmentation, object contour and location assumption.  Deep3DBox \cite{mousavian20173d} utilizes the constraint relationship between the predicted 2D boxes and the projected 3D boxes to calculate 3D parameters of targets. DeepMANTA \cite{chabot2017deep} and 3D-RCNN \cite{kundu20183d} propose to  match 2D object proposals and predefined 3D CAD models to gradually refine the 3D parameters. M3D-RPN \cite{brazil2019m3d} leverages a depth-aware network to generate 2D and 3D proposals simultaneously. Beyond these methods, there are also many advanced methods are introduced recent years \cite{li2019gs3d,brazil2019m3d,liu2019deep,manhardt2019RoI,qin2019monogrnet,simonelli2019disentangling}, we kindly refer readers to \cite{fan2021deep} for more information about them.  Beyond direct prediction, there are also many works propose to generate pseudo-LiDAR data from a single RGB image to conduct 3D object detection, called pseudo-LiDAR based. For example, PL-MONO \cite{wang2019pseudo} uses a depth estimator to generate a depth map by predicting the depth on each image pixel, and then projects it to pseudo point cloud, which is then sent to the existing 3D detector as LiDAR signals to predict the target boundary. Next, many other researchers \cite{weng2019monocular,qian2020end,ma2020rethinking,ye2020monocular} explore different depth estimation strategies, different fusion methods, different network architectures to try to improve the performance of pseudo-LiDAR based. Representatively, CaDDN \cite{reading2021categorical} performs both probabilistic depth estimation and training 3D detection in an end-to-end fashion, which to some extent eliminates the disadvantages of depth estimation mentioned in \cref{disadvantages of depth estimation}. However, the quality of additional spatial clues still depends on the depth distribution network branch, its accuracy is weaker than that of point cloud containing explicit spatial information. Nevertheless, CaDDN \cite{reading2021categorical} still achieves the state-of-the-art performance. Thanks to the additional  estimated depth information, these methods always perform better than those direct prediction methods.

Some recent researches attempt to extract robust 3D features from point clouds, depth maps or stereo images to enhance monocular feature extraction and target detection, such as MonoDistill \cite{Chong2022MonoDistillLS} and SGM3D \cite{SGM3D}, which are somewhat different from our proposed MonoSIM. Specifically, MonoDistill \cite{Chong2022MonoDistillLS} uses dense depth maps and RGB monocular images to train two identical detectors, and then enhances monocular spatial cues at corresponding network locations. SGM3D \cite{SGM3D} adopts a similar design. It estimates the depth of stereo and monocular images, uniformly converts them to BEV features, and then sends them to similar detection networks. In contrast, the above two methods are only heterogeneous in data, while MonoSIM is heterogeneous in both data and spatial structure.

% Due to better performance of pseudo-LiDAR based methods, in this work, we also adopt such technical roadmap. The main difference between our work and existing pseudo-LiDAR based methods is that they are trying to simulate the input of 3D object detectors, while we propose to simulate the feature learning behavior of the 3D object detectors.
%%%%%%%%%%%%%%%%%%%%%%%%%%%%%%%%%%%%%%%%%%%%%%%%%%%%%%%%%%%%%%%%%%%%%%%%%%%%%%%%%%%%%%%%%%%%%%%%%%%%%%%%%%
\section{Method}
\cref{figure2:Pipeline of MonoSIM} illustrates the architecture of MonoSIM. In our work, we assume both the point cloud based detector and the monocular detector can be spitted to three sub-components: the backbone scene-level feature extraction component, the RoI extraction component, and the prediction head with loss function, which consist of the full pipeline of the detectors feature learning. Therefore, to simulate the point cloud based detector, we design the scene-level simulation module, RoI-level simulation module and response-level simulation module.  Next, we introduce them in details.
 
%%%%%%%%%%%%%%%%%%%%%%%%%%%%%%%%
\subsection{Scene-Level Simulation Module}

\begin{figure}[t]
\centering
\setlength{\belowcaptionskip}{-0.4cm}
\includegraphics[scale=0.65]{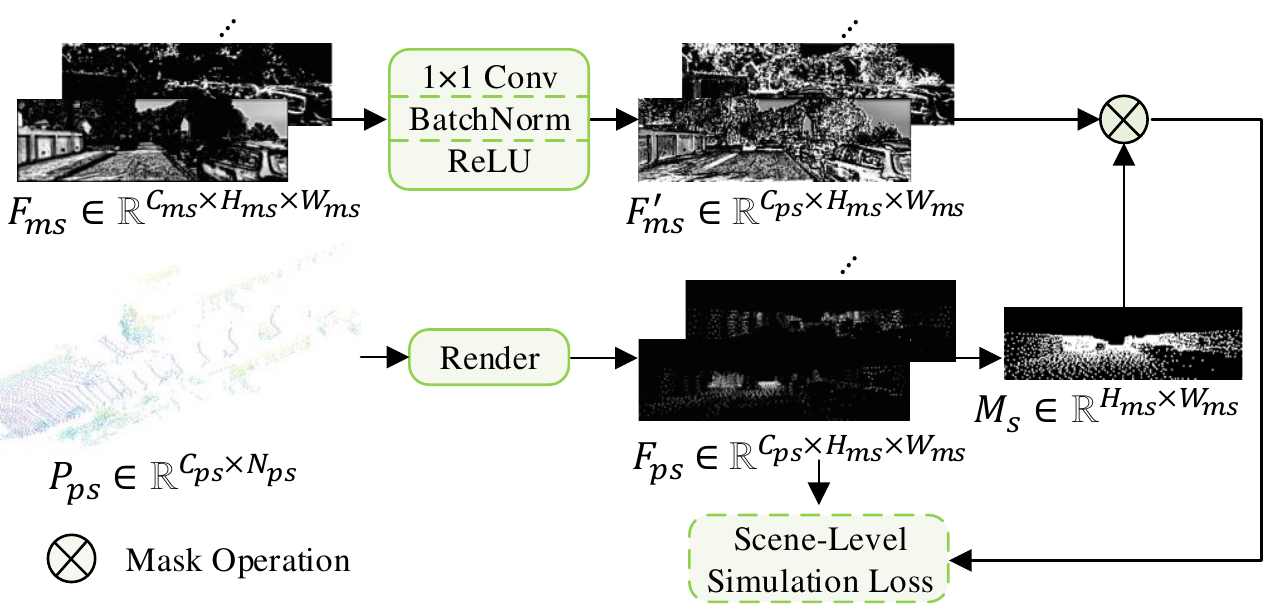}
\caption{Scene-Level Simulation Module. The scene-level branches are derived from the two modal detectors, the point cloud features needs to be converted into image features. The mask is used to filter the worthless zones, so that the monocular detector can simulate useful features.}
\label{figure3:Scene-Level Simulation Module.}
\end{figure}

The shallow scene-level features are the basis for detectors to perceive the environment, which are usually generated in the front part of networks, so we consider drawing scene-level branches from the backbone part for simulation. The scene-level simulation module is shown in \cref{figure3:Scene-Level Simulation Module.}.

% The visualization and details of scene-level simulation module are shown in \cref{figure3:Scene-Level Simulation Module.}.

\subsubsection{Scene-Level Feature Alignment}
Since the monocular and point cloud based detectors are completely different structures, there are cross modal differences in feature expression, forcing the monocular image network to directly simulate the spatial features of points will make the monocular network deviate from the correct optimization direction. Therefore, it is necessary to align the monocular and point cloud based detectors respectively to narrow the modal gap.

Monocular scene-level features are defined as $F_{ms} \in \mathbb{R} ^{C_{ms}\times H_{ms}\times W_{ms}}$, where $C_{ms}$ is the number of feature channels, $H_{ms}$, $W_{ms}$ are the height and width of the monocular scene features. Since monocular scene features are already image features, we only need to match the channels of monocular and point cloud scene features. Specifically, denoting the channels provided by the point cloud based detector is $C_{ps}$, we use a 1x1 Convolution + BatchNorm + ReLU layer to adjust $C_{ms}$ to $C_{ps}$. Therefore, the aligned monocular scene features $F_{ms}^{'} \in \mathbb{R} ^{C_{ps}\times H_{ms}\times W_{ms}}$ can be generated. 

We assume that the features of the scene point cloud is $P_{ps} \in \mathbb{R} ^{C_{ps}\times N_{ps}}$, where $N_{ps}$ is the number of scene feature points. Due to the modality differences, we need to convert point cloud features into image features. Specifically, we use render method to convert points into image.

The features of scene feature points can be expressed as $P_{ps} = {\{p^{ps}_{1},\cdots,p^{ps}_{N_{ps}}\}} \in \mathbb{R} ^{C_{ps}\times N_{ps}}$. $Q_{ps} = {\{q^{ps}_{1},\cdots,q^{ps}_{N_{ps}}\}} \in \mathbb{R} ^{3\times N_{ps}}$ represents the 3D space coordinates of scene feature points, internal parameter matrix of the camera coordinate system is $K$ and external parameter matrix is $RT$. We specify the height and width of the output features as $H_{ms}$ and $W_{ms}$. Using the PyTorch3D library \cite{ravi2020pytorch3d} to render the scene features $F_{ps} = {\{f^{ps}_{1},\cdots,f^{ps}_{C_{ps}}\}} \in \mathbb{R} ^{C_{ps}\times H_{ms}\times W_{ms}}$, and render operation can be defined as:
    \begin{equation}
    F_{ps}=Render(P_{ps},Q_{ps},K,RT,H_{ms},W_{ms})
    \end{equation}

However in some backgrounds or blind zones of the rendered features, the point cloud based detector cannot provide any effective spatial cues for simulation, where values are 0 in all channels. These zones should be deleted in the subsequent simulation, so it is necessary to generate a scene mask.Specifically, we denote the scene mask is $M_s{(u,v)} \in \mathbb{R} ^{H_{ms}\times W_{ms}}$, each channel of $F_{ps}$ is $f_i^{ps}{(u,v)}$, where $i\in [1,C_{ps}]$, $u\in [1,H_{ms}]$, $v\in [1,W_{ms}]$, $M_s{(u,v)}$ can be defined as:
    \begin{equation}
    M_s{(u,v)}=
    \begin{cases}
    0& \text{$\sum_{i=1}^{C_{ps}}f_i^{ps}{(u,v)}=0$}\\
    1& \text{$\sum_{i=1}^{C_{ps}}f_i^{ps}{(u,v)}\neq 0$}
    \end{cases}
    \end{equation}

\subsubsection{Scene-Level Simulation Loss}

After obtaining the scene mask, we hope that in the limited zones, the monocular detector can simulate the feature distribution of the point cloud based detector as much as possible. Specifically, we use L1 norm to design scene-level simulation loss function $L_{scene}$.
    \begin{equation}
    L_{scene}=\frac{1}{n_s}\Vert {M_s(F_{ms}^{'}-F_{ps})} \Vert_1
    \end{equation}
where $n_s$ is the number of valid scene features in $M_s$.

%%%%%%%%%%%%%%%%%%%%%%%%%%%%%%%%
\subsection{RoI-Level Simulation Module}

\begin{table*}
    \centering
    \setlength{\belowcaptionskip}{-0.4cm}
    \small
    \begin{tabular}{ccccccccc}
        \toprule[1.5pt]
        \multirow{2}{*}{Method}& \multirow{2}{*}{Extra Data}& \multicolumn{3}{c}{$AP_{{3D|R40}}$}& \multicolumn{3}{c}{BEV}& Runtime\\
        & & easy& moderate& hard& easy& moderate& hard& (ms)\\
        \hline
        AM3D \cite{ma2019accurate}& Depth& 16.50& 10.74& 9.52& 25.03& 17.32& 14.91& 400\\
        DA-3Ddet \cite{ye2020monocular}& Depth& 16.80& 11.50& 8.90& -& -& -& -\\
        % PatchNet \cite{ma2020rethinking}& Depth& 15.68& 11.12& 10.17& 22.97& 16.86& 14.97& 400 \\
        D4LCN \cite{ding2020learning}& Depth& 16.65& 11.72& 9.51& 22.51& 16.02& 12.55& 200\\
        Kinem3D \cite{brazil2020kinematic}& Multi-frames& 19.07& 12.72& 9.17& 26.69& 17.52& 13.10& 120\\
        CaDDN \cite{reading2021categorical}& Depth& 19.17& 13.41& 11.46& 27.94& 18.91& 17.19& 630\\
        DFR-Net \cite{zou2021devil}& Depth& 19.40& 13.63& 10.35& 28.17& 19.17& 14.84& 180\\
        DDMP-3D \cite{2021Depth}& Depth& 19.71& 12.78& 9.80& 28.08& 17.89& 13.44& 180\\
        SGM3D \cite{SGM3D}& Stereo& 22.46& 14.65& 12.97& 31.49& 21.37& 18.43& 30\\
        MonoDistill \cite{Chong2022MonoDistillLS}& Depth& 22.97& 16.03& 13.60& 31.87& 22.59& 19.72& 40\\
        \hline
        M3D-RPN \cite{brazil2019m3d}& -& 14.76& 9.71& 7.42& 21.02& 13.67& 10.23& 160\\
        % RTM3D \cite{2020RTM3D}& -& 14.41& 10.34& 8.77& 19.17& 14.20& 11.99& 50\\
        MonoDLE \cite{2021Delving}& -& 17.23& 12.26& 10.29& 24.79& 18.89& 16.00& 40\\
        MonoPair \cite{2020MonoPair}& -& 13.04& 9.99& 8.65& 19.28& 14.83& 12.89& 60\\
        MonoRUn \cite{2021MonoRUn}& -& 19.65& 12.30& 10.58& 27.94& 17.34& 15.24& 70\\
        GrooMeD \cite{2021GrooMeD}& -& 18.10& 12.32& 9.65& 26.19& 18.27& 14.05& 120\\
        MonoRCNN \cite{2021Geometry}& -& 18.36& 12.65& 10.03& 25.48& 18.11& 14.10& 70\\
        MonoFlex \cite{2021Flexible}& -& 19.94& 13.89& 12.07& 28.23& 19.75& 16.89& 30\\
        GUPNet \cite{2021GeometryUncertainty}& -& 20.11& 14.20& 11.77& -& -& -& 30\\
        \hline        
        MonoSIM (ours)& \multirow{2}{*}{Depth}& 20.31& 13.74& 12.31& 28.27& 19.89& 17.96& \multirow{2}{*}{140}\\
        \textit{Improvement on baseline}& & \textbf{+1.14}& \textbf{+0.33}& \textbf{+0.85}& \textbf{+0.33}& \textbf{+0.98}& \textbf{+0.77}& \\
        \bottomrule[1.5pt]
    \end{tabular}
    \caption{Performance comparison on the KITTI test set for Car category. MonoSIM uses CaDDN \cite{reading2021categorical} as the monocular baseline and depth as auxiliary data during training.}
    \label{tab:Results comparison on the KITTI test set}
\end{table*}

\begin{figure}[t]
\centering
\setlength{\belowcaptionskip}{-0.4cm}
\includegraphics[scale=0.65]{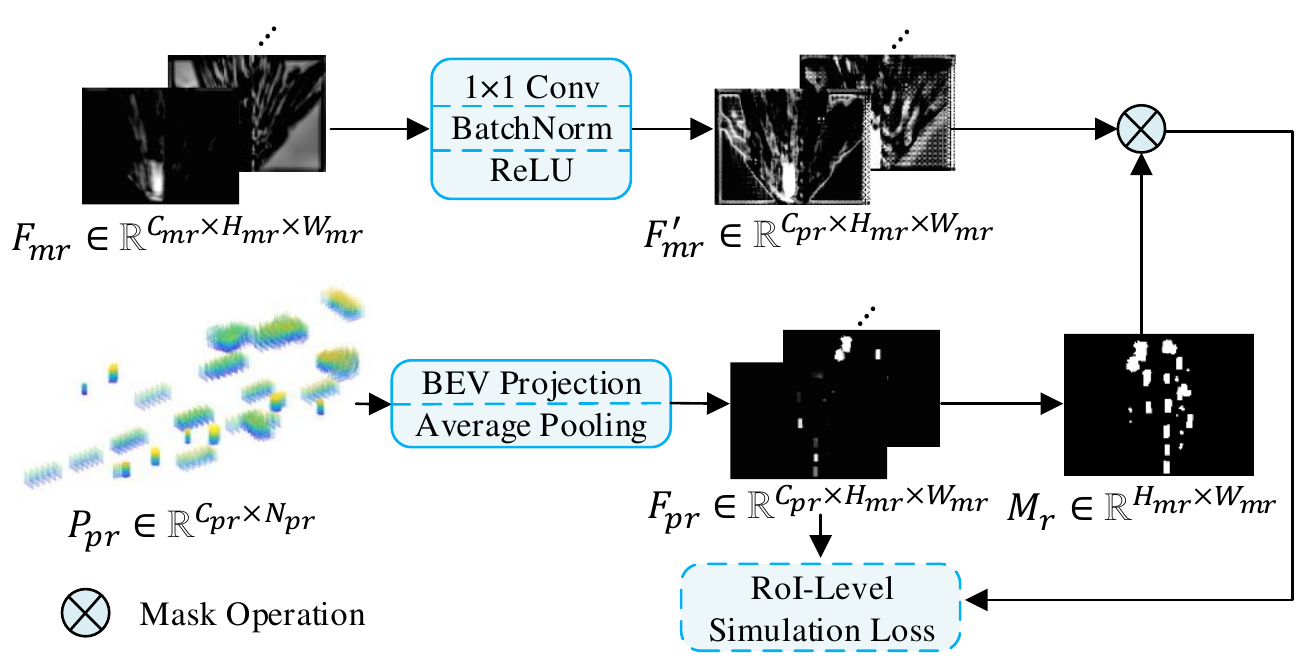}
\caption{RoI-Level Simulation Module. To be more flexible to different types of detectors, we propose another method of BEV projection to align RoI-level point cloud features.}
\label{figure4:RoI-Level Simulation Module.}
\end{figure}

Monocular detector simulates the generation of scene-level features and aims to improve the understanding ability of the environment. However in the whole environment, the target zones generally occupy less space, where the shallow scene cues is not dense enough. To refine the detection, the monocular detector needs centralized RoI-level cues for simulation at these places.
Different from the scene-level branches led from the front part of detectors, RoI-level features are mainly distributed in the middle or rear of the network. Therefore, the position and form of the RoI-level branches are flexible. 
The RoI-level simulation module is shown in \cref{figure4:RoI-Level Simulation Module.}.

% The visualization and details of RoI-level simulation module are shown in \cref{figure4:RoI-Level Simulation Module.}.

\subsubsection{RoI-Level Feature Alignment.}
The above scene-level feature alignment has introduced a method to align point cloud and image features. For the case that monocular RoI-level features are image type, we can also align the features by this method. Next, we propose another method to align BEV features in RoI-level feature alignment.

Monocular RoI-level features can be described as $F_{mr} \in \mathbb{R} ^{C_{mr}\times H_{mr}\times W_{mr}}$, where $C_{mr}$ is the number of feature channels, $H_{mr}$, $W_{mr}$ are the height and  width of the monocular RoI features. Similar to the scene feature alignment, we adopt 1x1 Convolution + BatchNorm + ReLU layer to adjust $C_{mr}$ to $C_{pr}$ and generate the aligned monocular RoI features $F_{mr}^{'} \in \mathbb{R} ^{C_{pr}\times H_{mr}\times W_{mr}}$, where $C_{pr}$ is number of feature channels of RoI feature points.

We denote the features of RoI points are $P_{pr} = {\{p^{pr}_{1},\cdots,p^{pr}_{N_{pr}}\}} \in \mathbb{R} ^{C_{pr}\times N_{pr}}$, where $N_{pr}$ is the number of RoI feature points. $Q_{pr} = {\{q^{pr}_{1},\cdots,q^{pr}_{N_{pr}}\}} \in \mathbb{R} ^{3\times N_{pr}}$ represents the 3D space coordinates of RoI feature points. RoI feature points are divided into voxels with features $V_{pr} \in \mathbb{R} ^{C_{pr}\times X\times Y\times Z}$, where $X$, $Y$, $Z$ are the size of voxels. The BEV features $B_{pr} \in \mathbb{R} ^{C_{pr}\times X\times Y}$ are obtained by averaging the voxel features at the same ${(x,y)}$, where $x\in [1,X]$, $y\in [1,Y]$. However, the size of $B_{pr}$ may be inconsistent with the aligned monocular RoI features $F_{mr}^{'}$, thus our work adds average pooling to $B_{pr}$ and generate point cloud RoI features $F_{pr} \in \mathbb{R} ^{C_{pr}\times H_{mr}\times W_{mr}}$.
After generating $F_{pr}$, RoI mask $M_r \in \mathbb{R} ^{H_{mr}\times W_{mr}}$ also needs to be generated as same as $M_s$.

\subsubsection{RoI-Level Simulation Loss.}
We use L1 norm to enforce the monocular detector to simulate the RoI-level feature provided by the point cloud based detector, and RoI-level simulation loss $L_{RoI}$ can be formulated as:
    \begin{equation}
    L_{RoI}=\frac{1}{n_r}\Vert {M_r(F_{mr}^{'}-F_{pr})} \Vert_1
    \end{equation}
where $n_r$ is the number of valid RoI features in $M_r$.

\begin{table*}
    \centering
    \setlength{\belowcaptionskip}{-0.4cm}
    \small
    \begin{tabular}{cccccccccc}
        \toprule[1.5pt]
        \multirow{2}{*}{Difficulty}& \multirow{2}{*}{Method}& \multicolumn{4}{c}{3D mAP}& \multicolumn{4}{c}{3D mAPH}\\
        & & overall& 0-30m& 30-50m& 50m-$\infty$& overall& 0-30m& 30-50m& 50m-$\infty$\\
        \hline
        \multirow{4}{*}{\makecell[c]{LEVEL\_1\\IoU=0.7}}& MonoDistill \cite{Chong2022MonoDistillLS}& 0.42& 1.23& 0.14& \textbf{0.03}& 0.25& 0.74& 0.08& \textbf{0.02}\\
        & M3D-RPN \cite{brazil2019m3d}& 0.35& 1.12& 0.18& 0.02& 0.34& 1.10& 0.18& \textbf{0.02}\\
        & MonoSIM (ours)& \textbf{1.60}& \textbf{6.08}& \textbf{0.40}& 0.01& \textbf{1.59}& \textbf{6.02}& \textbf{0.39}& 0.01\\
        & \textit{Improvement on baseline}& +1.25& +4.96& +0.22& -0.01& +1.25& +4.92& +0.21& -0.01\\
        \hline
        \multirow{4}{*}{\makecell[c]{LEVEL\_2\\IoU=0.7}}& MonoDistill \cite{Chong2022MonoDistillLS}& 0.40& 1.23& 0.13& \textbf{0.03}& 0.23& 0.73& 0.08& \textbf{0.02}\\
        & M3D-RPN \cite{brazil2019m3d}& 0.33& 1.12& 0.18& 0.02& 0.33& 1.10& 0.17& \textbf{0.02}\\
        & MonoSIM (ours)& \textbf{1.49}& \textbf{6.05}& \textbf{0.38}& 0.004& \textbf{1.48}& \textbf{6.00}& \textbf{0.38}& 0.004\\
        & \textit{Improvement on baseline}& +1.16& +4.93& +0.20& -0.016& +1.15& +4.90& +0.21& -0.016\\
        \hline
        \multirow{4}{*}{\makecell[c]{LEVEL\_1\\IoU=0.5}}& MonoDistill \cite{Chong2022MonoDistillLS}& 6.32& 12.56& \textbf{5.24}& \textbf{1.49}& 3.82& 7.35& \textbf{3.68}& \textbf{1.10} \\
        & M3D-RPN \cite{brazil2019m3d}& 3.79& 11.14& 2.16& 0.26& 3.63& 10.70& 2.09& 0.21\\
        & MonoSIM (ours)& \textbf{8.16}& \textbf{25.86}& 3.62& 0.08& \textbf{8.04}& \textbf{25.48}& 3.55& 0.08\\
        & \textit{Improvement on baseline}& +4.37& +14.72& +1.46& -0.18& +4.41& +14.78& +1.46& -0.13\\
        \hline
        \multirow{4}{*}{\makecell[c]{LEVEL\_2\\IoU=0.5}}& MonoDistill \cite{Chong2022MonoDistillLS}& 5.87& 12.50& \textbf{5.06}& \textbf{1.28}& 3.55& 7.32& \textbf{3.55}& \textbf{0.94}\\
        & M3D-RPN \cite{brazil2019m3d}& 3.61& 11.12& 2.12& 0.24& 3.46& 10.67& 2.04& 0.20\\
        & MonoSIM (ours)& \textbf{7.58}& \textbf{25.75}& 3.49& 0.07& \textbf{7.47}& \textbf{25.37}& 3.43& 0.07\\
        & \textit{Improvement on baseline}& +3.97& +14.63& +1.37& -0.17& +4.01& +14.70& +1.39& -0.13\\
        \toprule[1.5pt]
    \end{tabular}
    \caption{Performance comparison on the Waymo Open dataset for Vehicle category. MonoSIM uses M3D-RPN \cite{brazil2019m3d} as the monocular baseline. We sample the Waymo Open dataset according to \cref{dataset introduction} and retrain MonoDistill with the parameter settings provided by \cite{Chong2022MonoDistillLS}.}
    \label{tab:Results on the Waymo Open Dataset.}
\end{table*}

%%%%%%%%%%%%%%%%%%%%%%%%%%%%%%%%
\subsection{Response-Level Simulation Module}
To enhance the geometric parameter estimation of the object pose, we adopt the soft labels predicted by point cloud based detectors to supervise the monocular network training. 
% The response-level simulation module is shown in \cref{figure2:Pipeline of MonoSIM}.

Since the predicted soft labels have been completely aligned with the ground-truth labels in content and format, we directly replace the ground-truth labels with these soft labels. Response-level simulation loss $L_{response}$ can be expressed as:
    \begin{equation}
    L_{response}=L_{baseline}
    \end{equation}
where $L_{baseline}$ is defined by the monocular network.

\subsection{Total Simulation Loss}
The total loss of the MonoSIM is the combination of the above three parts:
    \begin{equation}
    L=L_{response}+{\lambda}_{scene} L_{scene}+{\lambda}_{RoI} L_{RoI}
    \label{equ:total loss of the MonoSIM.}
    \end{equation}
where ${\lambda}_{scene}$ and ${\lambda}_{RoI}$ are fixed loss weighting factors.

%%%%%%%%%%%%%%%%%%%%%%%%%%%%%%%%
\subsection{Application}
Note MonoSIM is model agnostic. Now, to verify its effectiveness, we apply it into current existing monocular 3D object detectors.  In our work, we decide to let the monocular detectors to simulate the behavior of PV-RCNN \cite{shi2020pv} due to PV-RCNN's strong power in 3D object detection and its wide usage. Then, we choose M3D-RPN \cite{brazil2019m3d} and  CaDDN \cite{reading2021categorical} as our baseline monocular detectors. The former is a classic anchor-based method, and the later is one of the current state-of-the-art BEV-based methods. 

%%%%%%%%%%%%%%%%%%%%%%%%%%%%%%%%%%%%%%%%%%%%%%%%%%%%%%%%%%%%%%%%%%%%%%%%%%%%%%%%%%%%%%%%%%%%%%%%%%%%%%%%%%
\section{Experiments}

\subsection{Dataset}
\label{dataset introduction}
To verify the effectiveness of our methods, we conduct experiments on the KITTI dataset \cite{Geiger2012CVPR}  and Waymo Open dataset \cite{Sun_2020_CVPR}.

\noindent \textbf{KITTI} dataset is one of the most widely used 3D detection datasets.  It contains 7481 training samples and 7518 test samples \cite{Geiger2012CVPR}. The training samples are divided into train set (3712 samples) and val set (3769 samples) following \cite{reading2021categorical}. We train the model and conduct ablation studies on the train set and val set, and submit the results of the test set to their lead board  to compare with the other existing state-of-the-art methods. Following previous methods, we only consider the "Car" category in KITTI dataset. 
 
\noindent \textbf{Waymo Open} dataset is the recently released large-scale autonomous driving 3D detection dataset, which consists of 798 training sequences, 202 validation sequences and 150 test sequences \cite{Sun_2020_CVPR}. Due to the large amount of data and high frame rate, we sample 20 percent of training sequences and val sequences to form train set (30926 samples) and val set (7839 samples). We detect vehicles, pedestrians and cyclists in Waymo annotations from images captured by the front camera.

\subsection{Implementation details}
We implement all our code using PyTorch. We use M3D-RPN \cite{brazil2019m3d} to conduct experiments on both Waymo Open dataset \cite{Sun_2020_CVPR} and KITTI dataset \cite{Geiger2012CVPR} and use CaDDN \cite{reading2021categorical} to conduct experiments on KITTI dataset. During training, we use \cref{equ:total loss of the MonoSIM.} to calculate loss with ${\lambda}_{scene}={\lambda}_{RoI}=1$ and remove the original flipping operation in data augment. For training CaDDN on the KITTI dataset, we adopt the Adam optimizer with a batch size 2. The learning rate  is 0.0002. The network is trained for 10 epochs. For training M3D-RPN, we adopt the SGD optimizer with a batch size 2. The learning rate is 0.004 with a poly decay rate using power 0.9 and eight decay of 0.9. The max iteration of the model is 200000. All experiments are conducted on a single  Tesla V100 (32G) GPU.

\subsection{Results on the KITTI Dataset}
%\subsubsection{Evaluation metrics.}
%To submit the detection results on test set and compare with other monocular methods on the official KITTI test server, average precision ($AP_{{3D|R40}}$) (40 recall positions) is adopted. The object categories include Car, Pedestrian and Cyclist. The Car IoU threshold is 0.7 and the IoU thresholds of Pedestrian and Cyclist are both 0.5.

We show the performance of our method in \cref{tab:Results comparison on the KITTI test set} and compare it with some state-of-the-art methods. The baseline method is CaDDN \cite{reading2021categorical}. The evaluation metric is the standard average precision ($AP_{{3D|R40}}$) (40 recall positions). In \cref{tab:Results comparison on the KITTI test set}, it can be obviously find that our MonoSIM improves our baseline model CaDDN for large margin while doesn't affect its original network architectures. Superficially,  the $AP_{{3D|R40}}$ is increased by 1.14\%, 0.33\% and 0.85\% on easy, moderate and hard difficulty levels, respectively. We contribute the improvement to our training pipeline: simulation on the feature learning behavior of existing point cloud based detectors. We can also find that our method outperforms most existing pseudo-LiDAR based methods. In contrast to simulate the input of point cloud based detectors, we simulate their feature learning behavior, so our method can learn stronger features, hence our performance is better.  

Note MonoDistill \cite{Chong2022MonoDistillLS} and SGM3D \cite{SGM3D} use same or similar model structures, which means that it is easier to align the modalities and scales of features, so as to obtain better training guidance. MonoSIM is committed to exploring a more flexible paradigm for monocular and point cloud simulation, so heterogeneous models are used to balance feature correspondence and method flexibility. Finally, \cref{tab:Results comparison on the KITTI test set} demonstrates that our method achieves performance close to that of state-of-the-art methods on the KITTI dataset.

% Moreover, by eliminating the depth estimation network, our method is much more efficient than existing pseudo-LiDAR based methods.

\subsection{Results on the Waymo Open Dataset}

\cref{tab:Results on the Waymo Open Dataset.} shows the performance of MonoSIM on Waymo Open dataset. 
% The baseline is M3D-RPN \cite{brazil2019m3d}. 
We adopt the official metrics: the mean average precision (mAP) and the mean average precision weighted by heading (mAPH) to evaluate the methods.  The evaluation levels are officially defines to two levels (LEVEL\_1, LEVEL\_2) according detection difficulty. IoU thresholds are set to 0.7 and 0.5 respectively to compute the metrics.  It can be found that our method improves the performance of the baseline on nearly all difficulty levels and evaluation thresholds. For instance, the overall mAP is improved by 1.25\% and 1.16\% at level 1 and level 2 respectively when the IoU threshold is 0.7, which is a significant improvement. When the IoU  threshold is 0.5, the improvement is more obvious, for level 1, it is increased from 3.79\% to 8.16\%, and for level 2, it is increased from 3.61\% to 7.58\%. Compared with the current state-of-the-art MonoDistill \cite{Chong2022MonoDistillLS}, MonoSIM improves the overall mAP by 1.18\% and 1.09\% at level 1 and level 2 respectively when the IoU threshold is 0.7.
The above results greatly demonstrate the effectiveness of MonoSIM.
% Note the superiority of MonoSIM comes from the simulation of the point cloud based detector's feature learning behavior.

\begin{table}
    \centering
    \setlength{\belowcaptionskip}{-0.4cm}
    \small
    \begin{tabular}{ccccc}
        \toprule[1.5pt]
        \multirow{2}{*}{Threshold}& {Number of}& \multicolumn{3}{c}{$AP_{{3D|R40}}$@IoU=0.7} \\
        & {Annotations}& easy& moderate& hard \\
        \hline
        Ground-Truth& 14357& 23.57& 16.31& 13.84 \\
        0& 21157& 21.13& 15.64& 13.98 \\
        0.3& 17805& 21.69& 13.50& 13.56 \\
        0.5& 16312& 20.49& 14.94& 13.41 \\
        0.7& 15043& \textbf{24.20}& \textbf{16.68}& \textbf{14.94} \\
        0.9& 12639& 22.79& 16.02& 14.29 \\
        \toprule[1.5pt]
    \end{tabular}
    \caption{Different confidence thresholds for filtering soft labels. Performance of $AP_{{3D|R40}}$ on the KITTI val set for Car category.}
    \label{tab:Different thresholds for filtering soft labels.}
\end{table}

\subsection{Comparison of filtering soft labels}
\cref{tab:Different thresholds for filtering soft labels.} shows the performance of MonoSIM based on CaDDN\cite{reading2021categorical} using different confidence thresholds to filter soft labels predicted by PV-RCNN\cite{shi2020pv}. When soft labels are filtered by lower confidence, the supervision signal contains more wrong or inaccurate classification and location information, which interferes the training of monocular detector, so it performs worse than using ground-truth labels. The performance is improved when the confidence threshold is set to 0.9, but it still does not reach the level of ground-truth labels due to few reserved annotations. When the threshold value is set to 0.7, the quality and quantity of supervision signals reach a good balance, the monocular detector achieves the best performance, thus we set 0.7 as the threshold to filter soft labels for Response-Level simulation, which are then used to supervise the training of the monocular detectors.

% \begin{figure*}[t]
% \centering
% \includegraphics[width=1\textwidth]{figure5}
% \caption{Qualitative results of MonoSIM. Samples are all from the KITTI val set on the Car category. \textbf{Red}, \textbf{blue} and \textbf{green} bounding boxes represent \textbf{ground-truth}, \textbf{baseline} and \textbf{MonoSIM}, separately. Depth refers to the distance of the object in front of the camera. In (a)-(f), baseline cannot perceive the objects while MonoSIM detects them correctly.}
% \label{figure5:Qualitative Results}
% \end{figure*}

\begin{figure*}[t]
\centering
\setlength{\belowcaptionskip}{-0.1cm}
  \begin{subfigure}{0.33\linewidth}
    \includegraphics[width=1\textwidth]{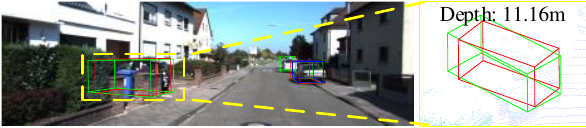}
    \caption{}
    \label{fig:5a}
  \end{subfigure}
  \hfill
  \begin{subfigure}{0.33\linewidth}
    \includegraphics[width=1\textwidth]{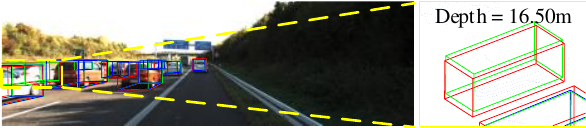}
    \caption{}
    \label{fig:5b}
  \end{subfigure}
  \hfill
  \begin{subfigure}{0.33\linewidth}
    \includegraphics[width=1\textwidth]{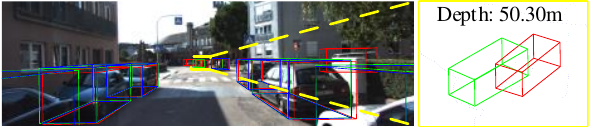}
    \caption{}
    \label{fig:5c}
  \end{subfigure}
  
  \begin{subfigure}{0.33\linewidth}
    \includegraphics[width=1\textwidth]{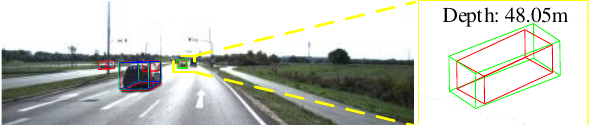}
    \caption{}
    \label{fig:5d}
  \end{subfigure}
  \hfill
  \begin{subfigure}{0.33\linewidth}
    \includegraphics[width=1\textwidth]{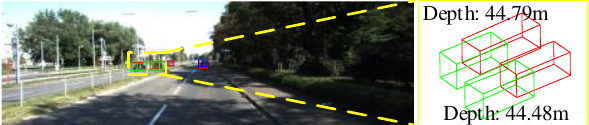}
    \caption{}
    \label{fig:5e}
  \end{subfigure}
  \hfill
  \begin{subfigure}{0.33\linewidth}
    \includegraphics[width=1\textwidth]{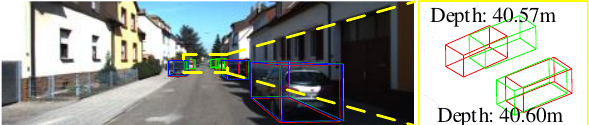}
    \caption{}
    \label{fig:5f}
  \end{subfigure}
  
  \setlength{\belowcaptionskip}{-0.2cm}
  \begin{subfigure}{0.33\linewidth}
    \includegraphics[width=1\textwidth]{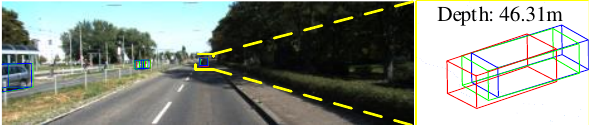}
    \caption{}
    \label{fig:5g}
  \end{subfigure}
  \hfill
  \begin{subfigure}{0.33\linewidth}
    \includegraphics[width=1\textwidth]{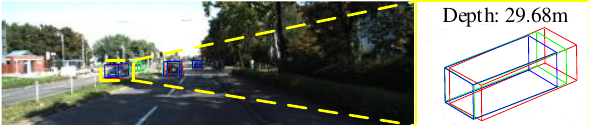}
    \caption{}
    \label{fig:5h}
  \end{subfigure}
  \hfill
  \begin{subfigure}{0.33\linewidth}
    \includegraphics[width=1\textwidth]{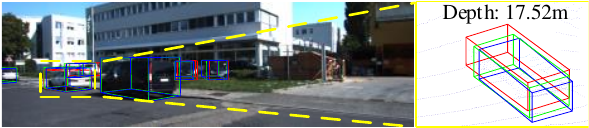}
    \caption{}
    \label{fig:5i}
  \end{subfigure}

\caption{Qualitative results of MonoSIM. Samples are all from the KITTI val set on the Car category. \textbf{Red}, \textbf{blue} and \textbf{green} bounding boxes represent \textbf{ground-truth}, \textbf{baseline} and \textbf{MonoSIM}, separately. Depth refers to the distance of the object in front of the camera. In \cref{fig:5a,fig:5b,fig:5c,fig:5d,fig:5e,fig:5f}, baseline cannot perceive the objects while MonoSIM detects them correctly.}
\label{figure5:Qualitative Results}
\end{figure*}

\subsection{Ablation Studies}
In this section, we conduct ablation studies on the KITTI val set to verify the effectiveness of our three simulation modules.  Both CaDDN \cite{reading2021categorical} and M3D-RPN \cite{brazil2019m3d} are used as our baselines. \cref{tab:Ablation studies of CaDDN simulation.} and \cref{tab:Ablation studies of M3D-RPN simulation(AP R40).} show the results. In the tables, SFS refers to the scene-level simulation module, RFS refers to the RoI-level simulation module, and RLS refers to the response-level simulation module.

\begin{table}
    \centering
    \small
    \begin{tabular}{ccccccc}
        \toprule[1.5pt]
        \multirow{2}{*}{Exp.}& \multicolumn{3}{c}{Method}& \multicolumn{3}{c}{$AP_{{3D|R40}}$@IoU=0.7} \\
        & RLS& SFS& RFS& easy& moderate& hard \\
        \hline
        1& & & & 23.57& 16.31& 13.84 \\
        2& \Checkmark& & & 24.20& 16.68& 14.94 \\
        % & & \Checkmark& & 23.87& 16.49& 14.02 \\
        % & & & \Checkmark& 23.40& 16.05& 13.67 \\
        3& \Checkmark& \Checkmark& & 24.73& 16.68& 14.94 \\
        4& \Checkmark& & \Checkmark& 24.77& \textbf{17.29}& 14.77 \\
        5& \Checkmark& \Checkmark& \Checkmark& \textbf{25.13}& 16.98& \textbf{15.05} \\
        \toprule[1.5pt]
    \end{tabular}
    \caption{Ablation studies of MonoSIM based on CaDDN. Performance of $AP_{{3D|R40}}$ on the KITTI val set for Car category.}
    \label{tab:Ablation studies of CaDDN simulation.}
\end{table}

% \begin{table}
%     \centering
%     \small
%     \begin{tabular}{ccccccc}
%         \toprule[1.5pt]
%         \multirow{2}{*}{Exp.}& \multicolumn{3}{c}{Method}& \multicolumn{3}{c}{$AP_{3D}$@IoU=0.7} \\
%         & RLS& SFS& RFS& easy& moderate& hard \\
%         \hline
%         6& & & & 20.27& 17.06& 15.21 \\
%         7& \Checkmark& & & 24.07& 17.80& 16.34 \\
%         8& \Checkmark& \Checkmark& & 24.20& 17.99& 16.49 \\
%         9& \Checkmark& & \Checkmark& 25.03& 18.56& 17.04 \\
%         10& \Checkmark& \Checkmark& \Checkmark& \textbf{25.26}& \textbf{18.66}& \textbf{17.24} \\
%         \toprule[1.5pt]
%     \end{tabular}
%     \caption{MonoSIM based on M3D-RPN. Performance of $AP_{3D}$ for Car category.}
%     \label{tab:Ablation studies of M3D-RPN simulation.}
% \end{table}

\begin{table}
    \centering
    \setlength{\belowcaptionskip}{-0.4cm}
    \small
    \begin{tabular}{ccccccc}
        \toprule[1.5pt]
        \multirow{2}{*}{Exp.}& \multicolumn{3}{c}{Method}& \multicolumn{3}{c}{$AP_{{3D|R40}}$@IoU=0.7} \\
        & RLS& SFS& RFS& easy& moderate& hard \\
        \hline
        6& & & & 15.85& 11.52& 8.97 \\
        7& \Checkmark& & & 19.02& 12.94& 10.52 \\
        8& \Checkmark& \Checkmark& & 18.96& 12.96& 10.67 \\
        9& \Checkmark& & \Checkmark& 20.02& 13.71& 11.27 \\
        10& \Checkmark& \Checkmark& \Checkmark& \textbf{20.23}& \textbf{13.91}& \textbf{11.46} \\
        \toprule[1.5pt]
    \end{tabular}
    \caption{Ablation studies of MonoSIM based on M3D-RPN. Performance of $AP_{{3D|R40}}$ on the KITTI val set for Car category.}
    \label{tab:Ablation studies of M3D-RPN simulation(AP R40).}
\end{table}

\subsubsection{Effects of response-level simulation.}
In our work,  the response-level simulation module uses prediction of the point cloud based detector as soft labels to guide the loss computing step, aims at increasing the monocular detector’s ability of regressing geometric properties of the bounding box.  Experiment 2 and 7 shows that by adding this module, the $AP_{{3D|R40}}$ is increased by 0.63\%, 0.37\% and 1.1\% using CaDDN as baseline, and the $AP_{{3D|R40}}$ is increased by 3.17\%, 1.42\% and 1.55\%  using M3D-RPN as baseline. M3D-RPN's improvement on the easy level is very significant. That is because M3D-RPN is an early work with a relatively simple network architecture, which is hard for itself to learn strong features. So after using our response-level simulation module, the behavior of the point cloud based detector would greatly help it to make up for its disadvantages.

\subsubsection{Effects of scene-level simulation}
Experiments 3 and 8 shows the results of  adding scene-level simulation on the basis of response-level simulation module on the two monocular methods. The response-level simulation module aims at using prediction of the point cloud based detector to align the shallow scene-level features of the point cloud based and monocular detectors, so that increases the monocular detector’s environmental understanding ability. 
We find that this module also improve the performance of both baselines at all difficult levels, but the improvement is not significant. This is because this module works at the early stage of the feature learning process to help the model extract the basic environmental features better. These features are not directly corresponded to the objects, but they are also helpful.

\subsubsection{Effects of RoI-level simulation}
Experiment 4 and 9 show the results of adding RoI-level simulation module on the basis of response-level simulation module. 
The RoI-level simulation module simulates the feature characteristics of point cloud RoIs for each monocular RoI, aims at increasing the monocular detector’s ability of finding and locating the potential objects. 
It can be seen from the tables that after adding this module, the performance is improved by nearly 1\% for both baselines on the easy and moderate level, which greatly demonstrates the effectiveness of this module.

\subsubsection{Full performance on KITTI val set} 
Experiments 5 and 10 are our final version which adopts the full advantages of RLS, SFS and RFS.  Compared with baselines on the KITTI val set, our full MonoSIM improves the performance of the CaDDN detector by 1.56\%, 0.67\%, 1.21\%, and improves the performance of the M3D-RPN detector by 4.38\%, 2.39\%, 2.49\% on $AP_{{3D|R40}}$.

\subsection{Qualitative Analysis}
We choose CaDDN \cite{reading2021categorical} as the baseline.
As shown in \cref{fig:5a,fig:5b}, baseline is unable to detect the objects which are seriously occluded, while MonoSIM performs well.
% since MonoSIM obtains better detection ability against occlusion by simulating the point cloud based detector. 
When the target is too far and the baseline fails (see \cref{fig:5c,fig:5d,fig:5e,fig:5f}), MonoSIM still has a greater perceptive ability. Thanks to the spatial cues provided by distant point cloud, MonoSIM can improve the generation of features by learning the cues which are usually lost or incomplete in the image.
\cref{fig:5g,fig:5h,fig:5i} show that MonoSIM can also improve the detection performance at different distances, refining the 3D bounding boxes significantly.

%%%%%%%%%%%%%%%%%%%%%%%%%%%%%%%%%%%%%%%%%%%%%%%%%%%%%%%%%%%%%%%%%%%%%%%%%%%%%%%%%%%%%%%%%%%%%%%%%%%%%%%%%%
\section{Conclusion}

In this work, we propose MonoSIM, a novel monocular 3D object detection training pipeline, which aims at simulating the feature learning behaviors of strong point cloud based detectors. In MonoSIM, one scene-level simulation module, one RoI-level simulation module and one response-level simulation module are proposed to progressively simulate the full training pipeline of the point cloud based detector. Our method have been applied to the M3D-RPN detector and CaDDN detector. Experiments on KITTI dataset and Waymo Open dataset have demonstrated the effectiveness of our method. 

%%%%%%%%%%%%%%%%%%%%%%%%%%%%%%%%%%%%%%%%%%%%%%%%%%%%%%%%%%%%%%%%%%%%%%%%%%%%%%%%%%%%%%%%%%%%%%%%%%%%%%%%%%
%%%%%%%%% REFERENCES
{\small
\bibliographystyle{ieee_fullname}
\bibliography{MonoSIM}
}

%%%%%%%%% Supplementary 
%%%%%%%%% Main Body
\newpage

\begin{appendix}

\section{Simulation Branches Settings}

The structures of different modal detectors determine the different methods of dividing scene-level, RoI-level and response-level components. To avoid any changes on original networks, our method MonoSIM needs to draw scene-level, RoI-level and response-level branches from given detectors, considering their characteristics.

\begin{figure*}
\centering
\includegraphics[width=1\textwidth]{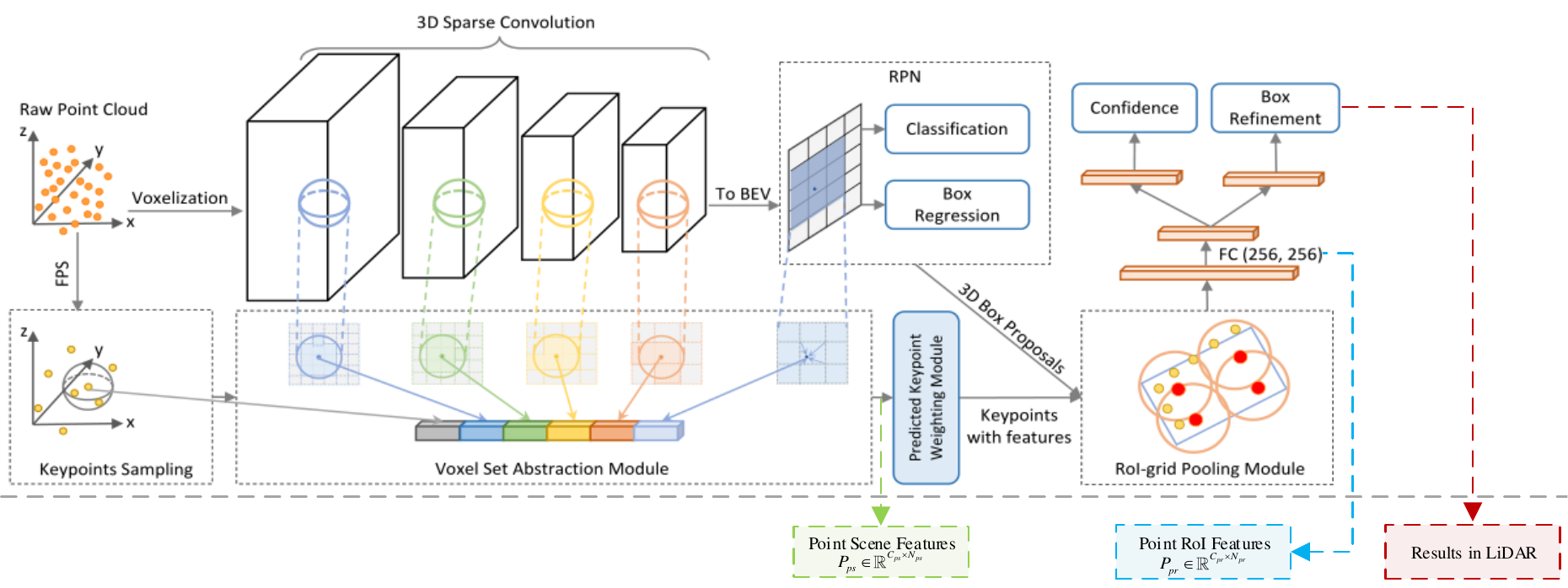}
\caption{The network pipeline above the gray dotted line is quoted from PV-RCNN \cite{shi2020pv} which is used as the point cloud based detector. Scene-level point features are extracted from the output of 'Voxel Set Abstraction Module'. RoI-level point features are extracted from the two-layer MLP after 'RoI-grid Pooling Module'.}
\label{supp_figure1:Branches extracted from PV-RCNN}
\end{figure*}

\begin{figure*}
\centering
\includegraphics[width=1\textwidth]{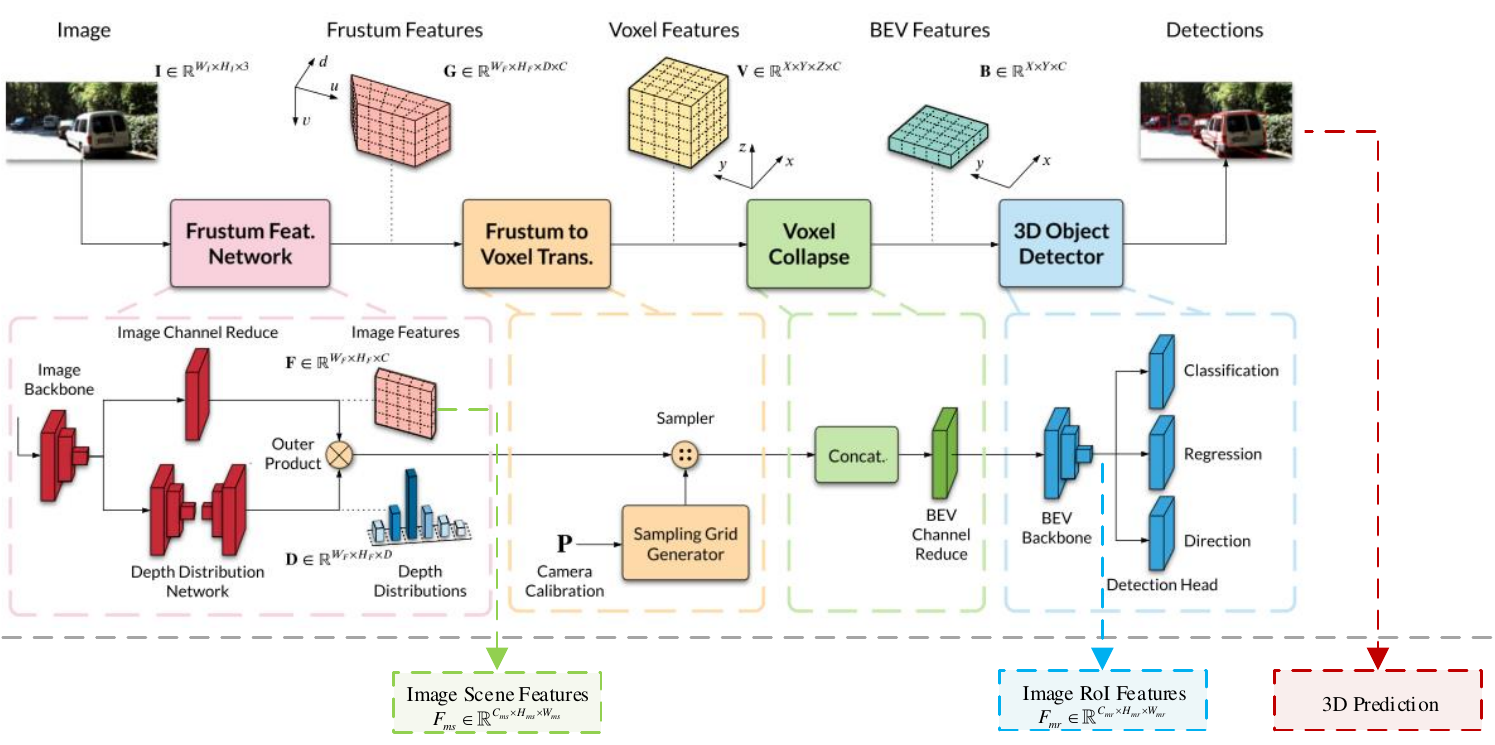}
\caption{The network pipeline above the gray dotted line is quoted from CaDDN \cite{reading2021categorical} which is used as the monocular detector. Scene-level image features are extracted from the output of 'Image Channel Reduce'. RoI-level image features are extracted from the output of the output of 'BEVBackbone'. }
\label{supp_figure2:Branches extracted from CaDDN}
\end{figure*}

\begin{figure*}
\centering
\includegraphics[width=1\textwidth]{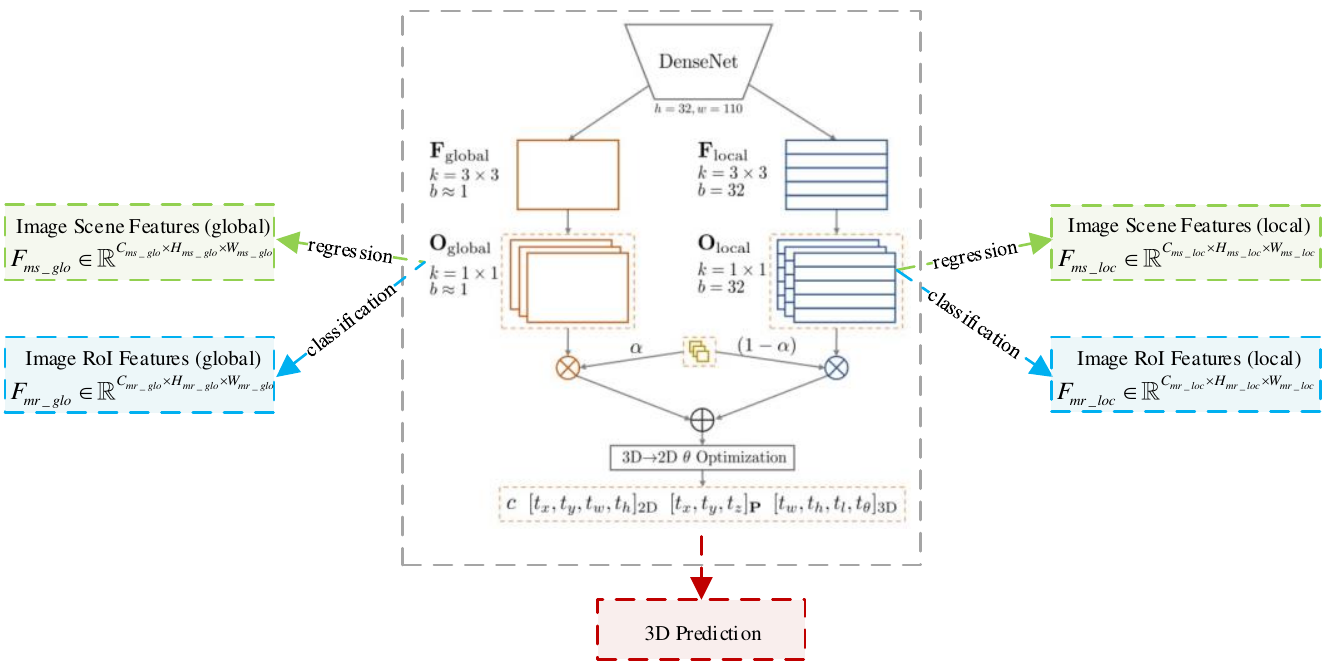}
\caption{ The network pipeline in the gray dotted box is quoted from M3D-RPN \cite{brazil2019m3d}. Due to the special structure of M3D-RPN, scene-level and RoI-level image features are all consisted of global and local branches. The scene and RoI features of each branch correspond to the regression and classification features of M3D-RPN, separately. }
\label{supp_figure3:Branches extracted from M3D-RPN}
\end{figure*}

\subsection{Point Cloud Based Detector}
We choose PV-RCNN \cite{shi2020pv} as the point cloud based detector, and the architecture of PV-RCNN is shown in \cref{supp_figure1:Branches extracted from PV-RCNN}. 

We think that the 'Voxel Set Abstraction Module' of PV-RCNN encodes the multi-scale features from 3D CNN which contains complete scene information, therefore we extract scene-level point features $P_{ps}$ from the output of 'Voxel Set Abstraction Module'. On the KITTI dataset, $P_{ps} \in \mathbb{R} ^{640\times 2048}$. On the Waymo Open dataset, $P_{ps} \in \mathbb{R} ^{544\times 4096}$. 

PV-RCNN is a two-stage detection framework and adopts RPN network to generate 3D proposals for RoI grids, this means  the RoI grids are list as potential target areas with centralized features. Therefore, RoI-level point features $P_{pr}$ are extracted from the output of the two-layer MLP after 'RoI-grid Pooling Module'. On both of KITTI and Waymo Open datasets, $P_{pr} \in \mathbb{R} ^{256\times 21600}$.

\subsection{Monocular Detector}
\subsubsection{CaDDN Detector}
The pipeline of CaDDN \cite{reading2021categorical} is shown in \cref{supp_figure2:Branches extracted from CaDDN}. We use CaDDN to conduct experiments on the KITTI dataset.

CaDDN adopts ResNet101 \cite{He_2016_CVPR} as the image backbone to acquire shallow environmental features, and then fused the channel reduced features and estimated depth information. Therefore, we extract features after 'Image Channel Reduce' as the image scene features $F_{ms} \in \mathbb{R} ^{64\times 94\times 311}$.

For RoI-level simulation, we hope that monocular features can approach the distribution of point RoI features as possible before entering the detection head. CaDDN's detection head takes BEV features as the input, so we branch out the RoI-level features $F_{mr}$ from here.  $F_{mr} \in \mathbb{R} ^{384\times 188\times 140}$.

\subsubsection{M3D-RPN Detector}
 M3D-RPN \cite{brazil2019m3d} is a one-stage detector leveraging the geometric relationship of 2D and 3D perspectives, without any extra components, thus the selection of scene-level and RoI-level branches are quite different from the CaDDN detector. The pipeline of M3D-RPN is shown in \cref{supp_figure3:Branches extracted from M3D-RPN}.
 
 M3D-RPN designs two parallel paths referred to global and local which are connected with the end of image backbone. The features,  which are sent into each path, are convoluted by each proposal feature extraction layer to generate new features ($F_{global}$ and $F_{local}$). $F_{global}$ and $F_{local}$ are then connected to kernels which can be divided as classification and regression branches. Considering point RoI features $P_{pr}$ which contains classification information from PV-RCNN, we extract features from the classification branch as image RoI features, and features from the regression branch as image scene features. 
 
 On global path, we define image scene and RoI features as $F_{ms\_glo} \in \mathbb{R} ^{C_{ms\_glo}\times H_{ms\_glo}\times W_{ms\_glo}}$ and $F_{mr\_glo} \in \mathbb{R} ^{C_{mr\_glo}\times H_{mr\_glo}\times W_{mr\_glo}}$, separately. On local path, we define image scene and RoI features as $F_{ms\_loc} \in \mathbb{R} ^{C_{ms\_loc}\times H_{ms\_loc}\times W_{ms\_loc}}$ and $F_{mr\_loc} \in \mathbb{R} ^{C_{mr\_loc}\times H_{mr\_loc}\times W_{mr\_loc}}$ separately. $C$, $H$ and $W$ represent the channel number, height and width of the corresponding features respectively.
 
 The reason for this definition is to follow the design of M3D-RPN which splits into global and local paths. We send  $F_{ms\_glo}$ and $F_{ms\_loc}$ into scene-level simulation module to simulate with $P_{ps}$ at the same time. Therefore, we can get scene-level simulation loss of $F_{ms\_glo}$ and $F_{ms\_loc}$ which are denoted as $L_{scene\_glo}$ and $L_{scene\_loc}$. In the same way, we can obtain RoI-level simulation loss of $F_{mr\_glo}$ and $F_{mr\_loc}$ which are denoted as $L_{RoI\_glo}$ and $L_{RoI\_loc}$.
 
 To leverage the global and local simulation, we use two learned weighting factors to fuse them:
    \begin{equation}
    L_{scene}=\alpha \cdot L_{scene\_glo} + (1-\alpha) \cdot L_{scene\_loc}
    \end{equation}
    \begin{equation}
    L_{RoI}=\beta \cdot L_{RoI\_glo} + (1-\beta) \cdot L_{RoI\_loc}
    \end{equation}
where $\alpha$ and $\beta$ are learnable parameters after sigmoid.

On the KITTI dataset, $F_{ms\_glo} \in \mathbb{R} ^{512\times 32\times 110}$, $F_{mr\_glo} \in \mathbb{R} ^{512\times 32\times 110}$, $F_{ms\_loc} \in \mathbb{R} ^{512\times 32\times 110}$ and $F_{mr\_loc} \in \mathbb{R} ^{512\times 32\times 110}$. On the Waymo Open dataset,  $F_{ms\_glo} \in \mathbb{R} ^{512\times 32\times 48}$, $F_{mr\_glo} \in \mathbb{R} ^{512\times 32\times 48}$, $F_{ms\_loc} \in \mathbb{R} ^{512\times 32\times 48}$ and $F_{mr\_loc} \in \mathbb{R} ^{512\times 32\times 48}$.

%%%%%%%%%%%%%%%%%%%%%%%%%%%%%%%%%%%%%%%%%%%%%%%%%%%%%%%%%%%%%%%%%%%%%%%%
\section{Soft Labels Settings}

\begin{figure}
\centering
\includegraphics[scale=0.7]{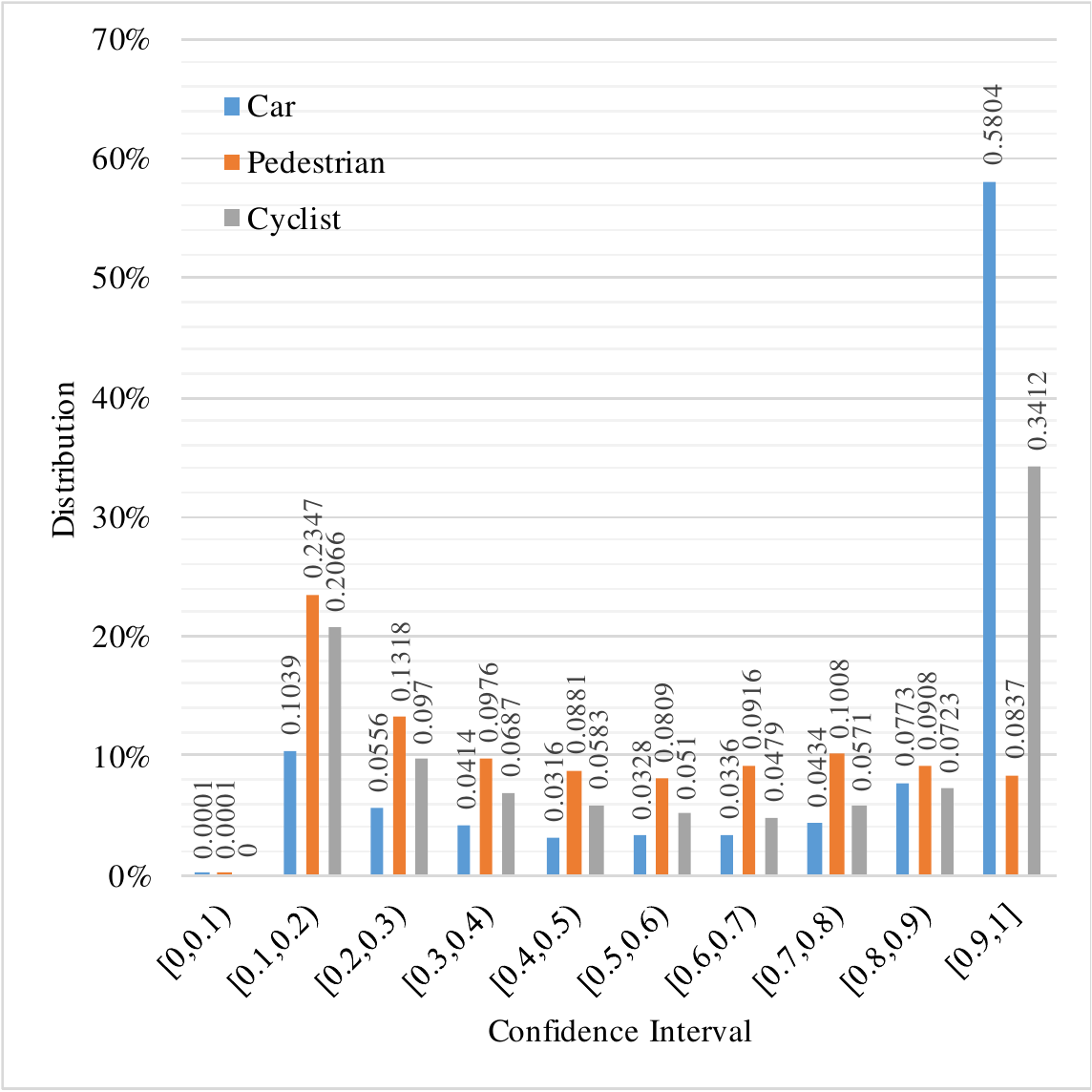}
\caption{Confidence distribution of PV-RCNN's prediction results on the KITTI train and val sets. }
\label{supp_figure4:Confidence distribution}
\end{figure}

In \cref{supp_figure4:Confidence distribution}, the confidence distribution of results predicted by PV-RCNN are different on Car, Pedestrian and Cyclist categories. For Car, prediction results with low confidence are few, and when the confidence reaches about 0.7, the proportion of high confidence results begins to rise rapidly. High confidence indicates more accurate detection, which also prove the high performance of PV-RCNN. For Pedestrian, the distribution is relatively uniform due to the difficulty of detecting this kind of objects. For Cyclist, the confidence distribution fluctuates obviously, but the samples are too insufficient.

Therefore, we filter PV-RCNN's inference results by confidence as the soft labels to supervise monocular detectors training. Specifically, the confidence threshold for Car category is 0.7 and set 0 to Pedestrian and Cyclist categories. In \cref{supp_tab:Filtering soft labels by confidence thresholds}, we compare the number of filtered samples and the ground-truth samples on the KITTI train and val sets. These filtered samples form soft labels for the response-level simulation module.

\begin{table*}
    \centering
    \small
    \begin{tabular}{ccccccc}
        \toprule[1.5pt]
        &  \multicolumn{3}{c}{Train Set}& \multicolumn{3}{c}{Val Set}\\
        & Car& Pedestrian& Cyclist& Car& Pedestrian& Cyclist\\
        \hline
        Ground-Truth Samples& 14357& 2207& 734& 14385& 2280& 893\\
        Filtered Samples& 15043& 7573& 1626& 16274& 9627& 1651\\

        \bottomrule[1.5pt]
    \end{tabular}
    \caption{Numbers of samples filtered by confidence thresholds on the KITTI train and val sets.}
    \label{supp_tab:Filtering soft labels by confidence thresholds}
\end{table*}

\begin{table*}
    \centering
    \small
    \begin{tabular}{ccccccc}
        \toprule[1.5pt]
        \multirow{2}{*}{Method}&  \multicolumn{3}{c}{Pedestrian $AP_{{3D|R40}}$}& \multicolumn{3}{c}{Cyclist $AP_{{3D|R40}}$}\\
        & easy& moderate& hard& easy& moderate& hard\\
        \hline
        M3D-RPN \cite{brazil2019m3d}& 4.92& 3.48& 2.94& 0.94& 0.65& 0.47\\
        CaDDN \cite{reading2021categorical}& 12.87& 8.14& 6.76& \textbf{7.00}& \textbf{3.14}& \textbf{3.30}\\
        DDMP-3D \cite{2021Depth}& 4.93& 3.55& 3.01& 4.18& 2.50& 2.32\\
        D4LCN \cite{ding2020learning}& 4.55& 3.42& 2.83& 2.45& 1.67& 1.36\\
        MonoDLE \cite{2021Delving}& 9.64& 6.55& 5.44& 4.59& 2.66& 2.45\\
        MonoPair \cite{2020MonoPair}& 10.02& 6.68& 5.53& 3.79& 2.21& 1.83\\
        MonoRUn \cite{2021MonoRUn}& 10.88& 6.78& 5.83& 1.01& 0.61& 0.48\\
        MonoFlex \cite{9578273}& 9.43& 6.31& 5.26& 4.17& 2.35& 2.04\\
        SGM3D \cite{SGM3D}& 13.99& \textbf{8.81}& 7.26& 5.49& 2.92& 2.64\\
        MonoDistill \cite{Chong2022MonoDistillLS}& 12.79& 8.17& \textbf{7.45}& 5.53& 2.81& 2.40\\
        \hline        
        MonoSIM (ours)& \textbf{14.16}& 8.63& 7.30& 4.07& 2.09& 2.12\\
        % \textit{Improvement on baseline}& -& +1.14& +0.33& +0.85& +0.33& +0.98& +0.77& -\\
        \bottomrule[1.5pt]
    \end{tabular}
    \caption{Results comparison on the KITTI test set for the Pedestrian and Cyclist categories. MonoSIM uses CaDDN \cite{reading2021categorical} as the monocular baseline.}
    \label{supp_tab:Results comparison on the KITTI test set}
\end{table*}

\begin{table}
    \centering
    \small
    \begin{tabular}{ccccccc}
        \toprule[1.5pt]
        \multirow{2}{*}{Exp.}& \multicolumn{3}{c}{Method}& \multicolumn{3}{c}{$AP_{3D}$@IoU=0.7} \\
        & RLS& SFS& RFS& easy& moderate& hard \\
        \hline
        1& & & & 20.27& 17.06& 15.21 \\
        2& \Checkmark& & & 24.07& 17.80& 16.34 \\
        3& \Checkmark& \Checkmark& & 24.20& 17.99& 16.49 \\
        4& \Checkmark& & \Checkmark& 25.03& 18.56& 17.04 \\
        5& \Checkmark& \Checkmark& \Checkmark& \textbf{25.26}& \textbf{18.66}& \textbf{17.24} \\
        \toprule[1.5pt]
    \end{tabular}
    \caption{MonoSIM based on M3D-RPN. Performance of $AP_{3D}$ for Car category.}
    \label{tab:Ablation studies of M3D-RPN simulation.}
\end{table}

%%%%%%%%%%%%%%%%%%%%%%%%%%%%%%%%%%%%%%%%%%%%%%%%%%%%%%%%%%%%%%%%%%%%%%%%
\section{Pedestrian/Cyclist Detection}
The detection performance of Pedestrian and Cyclist on the KITTI test set is shown in \cref{supp_tab:Results comparison on the KITTI test set}. 

Compared with baseline, MonoSIM achieves state-of-the-art performance for Pedestrian category which increases $AP_{{3D|R40}}$ by 1.29\%, 0.49\% and 0.54\% on easy, moderate and hard levels. However, there is some performance degradation for Cyclist category. We believe that this is due to insufficient training samples. As the number of samples compared in \cref{supp_tab:Filtering soft labels by confidence thresholds}, the filtered samples belonging to Cyclist are quite insufficient. In contrast, the number of Pedestrian samples has been increased significantly after filtering, and which is one of the reasons for performance improvement.

%%%%%%%%%%%%%%%%%%%%%%%%%%%%%%%%%%%%%%%%%%%%%%%%%%%%%%%%%%%%%%%%%%%%%%%%%%%%%%%%5
\section{Supplementary Description of Metrics}
On the KITTI dataset, Average Precision ($AP|_R$) is adopted as the metric for 3D detection which can be expressed  as:
    \begin{equation}
    AP|_R = \frac{1}{|R|}\sum_{r\in R}\rho_{interp} (r)
    \end{equation}
    where $\rho_{interp} (r)$ is the interpolation function which gives the precision value at recall $r$.

Before October 8, 2019, KITTI officially adopted \cite{2010ThePascal} eleven recall levels $r \in R_{11} = \{0,0.1,0.2,\cdots,1\}$ which used in the 3D detection metric $AP_{3D}$. In some other works, it is also called $AP_{{3D|R11}}$. However, this approach may include false precision improvements caused by recall starting at 0. 

$AP_{{3D|R40}}$ is proposed by \cite{9010618} to fix this problem. The solution is replacing $R_{11}$ with $R_{40} = \{1/40,2/40,3/40,\cdots,1\}$ that calculating $AP|_R$ on 40 recall positions without 0. KITTI followed the suggestions and updated the metric.

Since M3D-RPN \cite{brazil2019m3d} is an early published work, $AP_{3D}$ ($AP_{{3D|R11}}$) was used in their paper. To have a more intuitive comparison with the results in \cite{brazil2019m3d}, we uniformly use $AP_{3D}$ to conduct ablation studies on the KITTI val set as shown in \cref{tab:Ablation studies of M3D-RPN simulation.}. The experimental results show that MonoSIM can still significantly improve the detection performance of baseline under the evaluation Metrics of $AP_{3D}$.

%%%%%%%%%%%%%%%%%%%%%%%%%%%%%%%%%%%%%%%%%%%%%%%%%%%%%%%%%%%%%%%%%%%%%%%
\section{More Qualitative Results}
\begin{figure*}[t]
\centering
  \begin{subfigure}{1\linewidth}
    \includegraphics[width=1\textwidth]{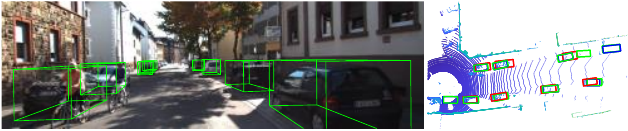}
    \caption{}
    \label{fig:supp_5a}
  \end{subfigure}

  \begin{subfigure}{1\linewidth}
    \includegraphics[width=1\textwidth]{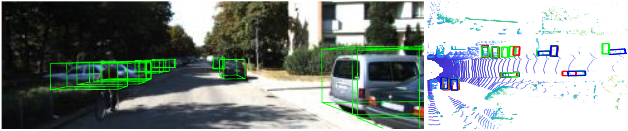}
    \caption{}
    \label{fig:supp_5b}
  \end{subfigure}

  \begin{subfigure}{1\linewidth}
    \includegraphics[width=1\textwidth]{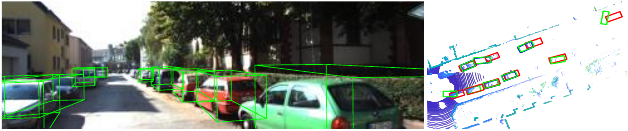}
    \caption{}
    \label{fig:supp_5c}
  \end{subfigure}
  
  \begin{subfigure}{1\linewidth}
    \includegraphics[width=1\textwidth]{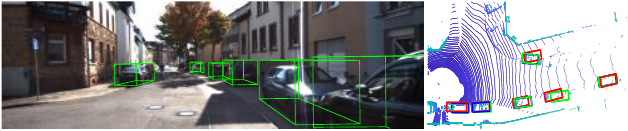}
    \caption{}
    \label{fig:supp_5d}
  \end{subfigure}

  \begin{subfigure}{1\linewidth}
    \includegraphics[width=1\textwidth]{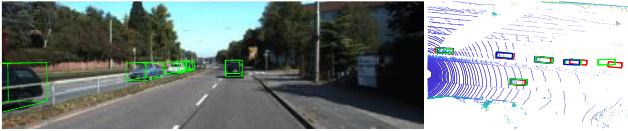}
    \caption{}
    \label{fig:supp_5e}
  \end{subfigure}

  \caption{We still use CaDDN as the baseline detector to conduct the qualitative comparison. \textbf{Red}, \textbf{blue} and \textbf{green} represent ground-truth, baseline and MonoSIM, separately. We only draw green bounding boxes in left images to show MonoSIM's detection.}
\label{supp_figure5:More Qualitative Results}
\end{figure*}

Since we have compared the methods by highlighting the bounding boxes of detected objects, now we qualitatively show the detection of all objects in the bird's-eye-view map. In \cref{supp_figure5:More Qualitative Results}, MonoSIM has achieved better detection results for multiple objects in different scenes, especially in the case of long distance and occlusion, the improvement is more significant.

\end{appendix}
%%%%%%%%%%%%%%%%

\end{document}